\pdfoutput=1

\documentclass[11pt]{article}

\usepackage{EMNLP2022}

\usepackage{times}
\usepackage{latexsym}

\usepackage{url}

\usepackage{hyperref}
\hypersetup{
    colorlinks=true,
    linkcolor=blue,
    filecolor=magenta,      
    urlcolor=blue,
    pdftitle={Don't Copy the Teacher: Data and Model Challenges in Embodied Dialogue},
    pdfpagemode=FullScreen,
    }
\hypersetup{citecolor=blue}
\usepackage[T1]{fontenc}

\usepackage[utf8]{inputenc}

\usepackage{microtype}

\usepackage{inconsolata}

\newcommand{\rowsqueeze}{\vspace{-2pt}}
\usepackage{tabularx}
\usepackage{multirow}
\usepackage{siunitx}
\usepackage{booktabs} 
\usepackage{enumitem}

\newcommand{\katerina}{Symbiote}
\newcommand{\Tiffany}[1]{\textcolor{black}{#1}}
\newcommand{\Tiffanyarxiv}[1]{\textcolor{black}{#1}}
\usepackage{graphicx} 

\title{Don't Copy the Teacher: \\
       Data and Model Challenges in Embodied Dialogue}

\author{So Yeon Min$^1$ \hspace{2em} Hao Zhu$^2$ \hspace{2em} Ruslan Salakhutdinov$^1$ \hspace{2em} Yonatan Bisk$^2$\\
  Machine Learning$^1$ and Language Technologies$^2$ at Carnegie Mellon University \\
  \texttt{\{soyeonm,hzhu2,rsalakhu,ybisk\}@andrew.cmu.edu} \\}

\begin{document}
\maketitle
\begin{abstract}
Embodied dialogue instruction following requires an agent to complete a complex sequence of tasks from a natural language exchange. The recent introduction of benchmarks \cite{padmakumar:teach} raises the question of how best to train and evaluate models for this multi-turn, multi-agent, long-horizon task. This paper contributes to that conversation, by arguing that imitation learning (IL) and related low-level metrics are actually misleading and do not align with the goals of embodied dialogue research and may hinder progress.

We provide empirical comparisons of metrics, analysis of three models, and make suggestions for how the field might best progress.
First, we observe that models trained with IL take spurious actions during evaluation.  
Second, we find that existing models fail to ground \textit{query} utterances, which are essential for task completion. Third, we argue evaluation should focus on higher-level semantic goals.
\footnote{Code to be released at \\ \url{https://github.com/soyeonm/TEACh_FILM}}

\end{abstract}

\section{Introduction}

Dialogue is key to how humans collaborate; through dialogue, we query information, confirm our understanding, or banter in a friendly manner. Since communication helps us work more efficiently and successfully, it is only natural to imbue for collaborative agents with this same ability. Most work has focused on grounded dialogues for embodied navigation \citep{thomason2020vision, justask, Roman2020} or limited interaction \cite{suhr2019executing}, which are narrower domains than the larger instruction following literature \citep{Tellex2011,Tellex2020, shridhar2020alfred,  quadcopter2, blukis2021persistent, min2021film}.

The first step towards engaging in a dialogue, is being able to understand and learn from it. Picture a child watching their parents with the goal to learn by imitation. They witness instructions, clarifications, mistakes, and banter.  Begging the question: \emph{What should  one learn from noisy natural dialogues?} 

Unlike in alinguistic tasks where modeling humans has recently proved helpful for search strategies \cite{Deitke2022}, we focus on language based tasks that require learning lexical-visual-action correspondences.
We discuss and compare three paradigms: Instruction Following (IF), actions from Entire Dialogue History (EDH) and Trajectory from Dialogue (TfD).  
The novel TEACh dataset \citep{TEACH} proposes EDH as the primary metric and 
uses the Episodic Transformer (ET) \citep{episodictransformer}  trained with behavior cloning as their baseline.
We also include comparisons to the EDH competitive \katerina\footnote{Model outputs provided by correspondence with the team.} system and we adapt FILM \cite{min2021film}, a recent method for general IF, to dialog instruction following (DIF) on TEACH.
FILM and \katerina{} belong to a different family of models, focusing on abstract planning trained at a higher semantic level than behavior cloning. This approach appears crucial for generalization and TfD evaluations. 

Most importantly, we analyze the human behaviors in TEACH and the corresponding effect on ET, \katerina, and FILM,
as representatives of existing model classes. 
From our findings, we suggest there are three major challenges the community must tackle to move forward in the nascent field of Dialogue based Instruction Following:

\paragraph{Recognizing mistakes}
Behavior cloning encourages replication of \Tiffanyarxiv{low-level} errors, \Tiffanyarxiv{but not high-level intentions.} Agents should learn \Tiffanyarxiv{to construe high-level intentions of demonstrations} and to deviate from demonstration \Tiffanyarxiv{errors}.

\paragraph{Grounding queries}
No approaches correctly ground \textit{``queries"} requesting information.

\paragraph{Evaluation}
Agent evaluation should focus on achieving goals rather than immitating procedures.

\begin{figure*}[!ht]
    \centering
    \includegraphics[width=0.9\textwidth, height=220pt]{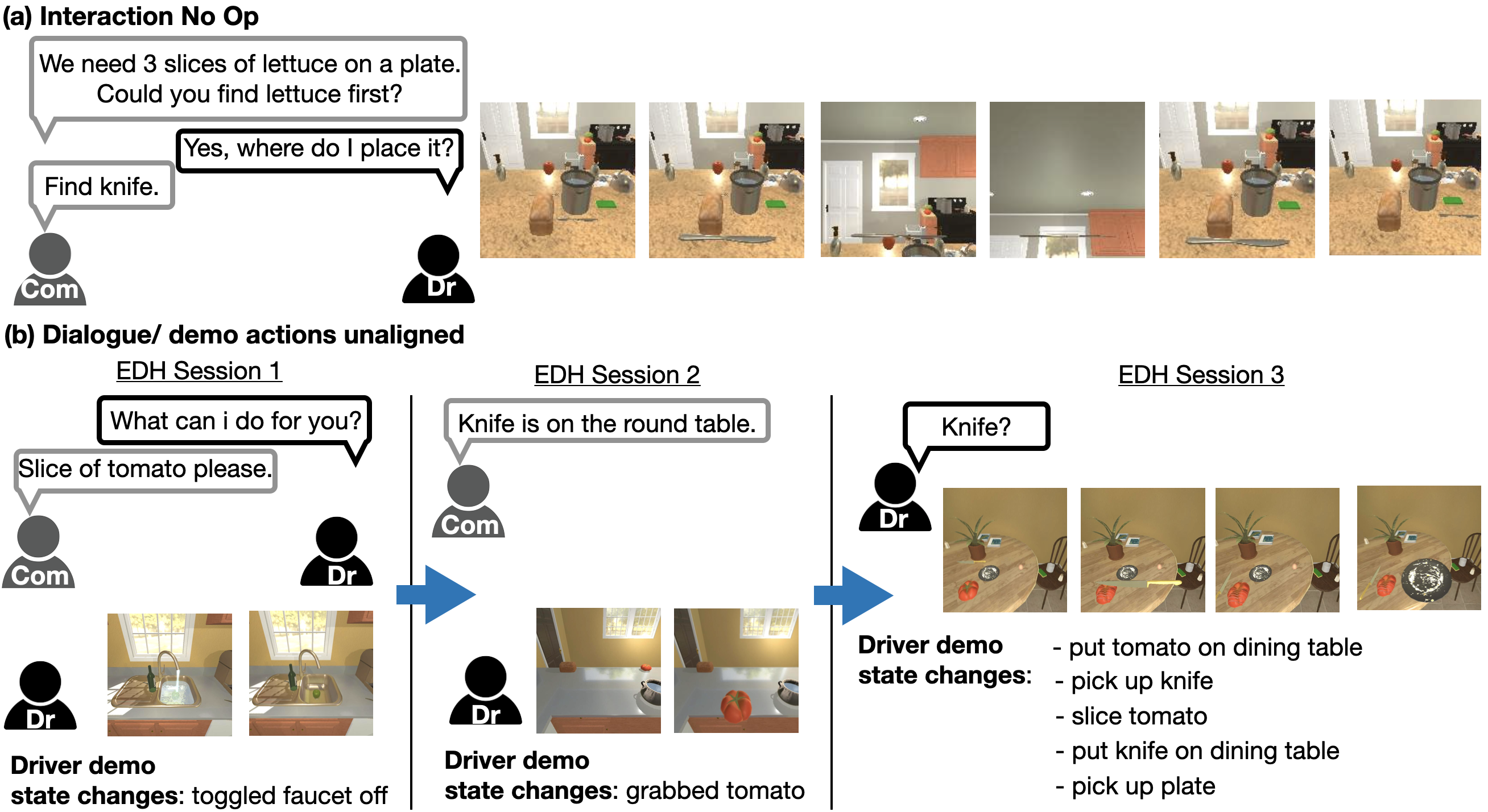}
    \caption{Examples of suboptimal demonstrations that can be harmful for training and evaluation. 
    (a: \textbf{no-op}) The driver grabs a knife, looks up and down, and put its down, although nowhere in the dialogue indicates to do these actions, nor do they facilitate the high-level goal. (b: \textbf{unaligned} intent) In EDH sessions 1 and 2, the commander asks for an item (a slice of tomato) and provides the location of the knife, but the driver performs unaligned actions. In session 3, the driver suddenly asks "knife?", but performs a long sequence of implied but not stated actions.}
    \label{fig:demo}
\end{figure*}

\section{Related Work}

\paragraph{Instruction Following}
A plethora of works have been introduced for instruction following without dialogue \cite{Chen2011,MatuszekISER2012}; an agent is expected to perform a task given a language instruction at the beginning and visual inputs at every time step. Representative tasks are Visual Language Navigation \citep{anderson2018vision, fried2018speaker, zhu2020vision} and instruction following (IF) \citep{shridhar2020alfred, singh2020moca}, which demands both navigation and manipulation. Popular methods rely on imitation learning \citep{episodictransformer, singh2020moca} and modularly trained components \citep{blukis2021persistent, min2021film} (e.g. for mapping and depth).

\paragraph{Dialogue Instruction Following }

Instruction Following with Dialogue \cite{she-etal-2014-back} has mostly addressed navigation. \citet{thomason2020vision,suhr2019executing} built navigation agents that ground human-human dialogues, while \citet{justask, nguyen-daume-iii-2019-help} showed that obtaining clarification via simulated interactions can improve navigation. 
Manipulation introduces 
grounding query utterances that involve more complex reasoning than in navigation-only scenarios \cite{Tellex2013}; for example, the agent may hear that the object of interest (e.g. ``apple'') is inside ``the third cabinet to the right of the fridge.''

\paragraph{Imitation Learning vs Higher semantics}
While behavior cloning (BC) is a popular method used to train IF agents, it assumes that expert demonstration is optimal \citep{IL_CAIL, wu2019imitation}. TEACh demonstrations are more ``ecologically valid" \cite{Vries2020} but correspondingly suboptimal, frequently containing mistakes and unnecessary actions. Popular methods that deal with suboptimal demonstrations involve annotated scoring labels or rankings for the quality of demonstrations \citep{wu2019imitation, T-rex}. Such additional annotations are not available in existing IF and DIF benchmarks. In this work, we empirically demonstrate the effect of noisy demonstrations on an episodic trained with BC for DIF.

\section{Tasks}
TEACh focuses on two tasks: Entire Dialogue History and Trajectory from Dialogue.  Despite what the name implies, EDH is an evaluation over partial dialogues (e.g. from state $S_t$ begin execution to $S_T$).  TfD starts an agent at $S_0$ and asks for a complete task completion provided the full dialogue.

In both settings, the agent (driver) completes household tasks conditioned on text, egocentric RGB observations, and the current view. An instance of a dialogue will take the form of a command: \textit{Prepare coffee in a clean mug. Mugs are in the microwave.}, the agent response \textit{How many do I need?}, and commander's answer: \textit{One}, 
together with a sequence of RGB frames and actions that the agent performed during the dialogue. As in this example, the agent has to achieve multiple subtasks (e.g. find mug in the microwave, clean mug in the sink, turn on the coffee machine, etc) to succeed. 

In TfD, the full dialogue history is given, and the agents succeeds if it completes the full task itself (e.g. make coffee). In EDH, the dialog history is partitioned into ``sessions'' (e.g. Fig. \ref{fig:demo}) 
with the corresponding action/vision/dialogue history until the first utterance of the commander (\textit{Prepare $\sim$ microwave.}) being the first session and those after it being the second. In EDH evaluation, the agent takes one session as input and predicts actions until the next session. An agent succeeds if it realizes all state changes (e.g. \textit{Mug: picked up}) that the human annotator performed.
Succinctly, TfD measures the full dialogue 
while EDH evaluates subsequences.

\section{Models}
TEACh is an important new task for the community. We analyze the provided baseline (ET), retrofit the ALFRED FILM model, and requested outputs from the authors of \katerina on the EDH leaderboard.

ET is a transformer for direct sequence imitation approach, that produces low-level actions conditioned on the accumulated visual and linguistic contexts.
In contrast, FILM consists of four submodules - semantic mapping, language processing, semantic policy, and deterministic policy modules. For the adaptation, we refactored the original code of FILM to the TEACH API, retrained the learned components of the semantic mapping module for the change in height and camera horizon, and retrained/rewrote the language processing module to take a dialogue history 
as input. 
The language processing (LP) module of FILM maps an instruction 
to a \textit{task type} and instruction-specific \textit{arguments}. 
For TfD this maps a dialogue to a sequence of tasks, while for EDH only the subsequence is mapped to an immediate action. Symbiote is a competitive modular method for EDH whose language understanding component is designed for dialogues (\S\ref{app:symbiote}).

\section{Challenges of Human Traces}
\label{sec:5}
First we present how TEACh and, by extension, future embodied dialogue settings present novel training and evaluation challenges as the data, by virtue of its authenticity, includes substantial noise in both the training and evaluation (despite filtering by the authors \S\ref{app:filter}). See 
\S\ref{app:stats} for how statistics were computed, for those not explained in this section.

\subsection{Explanation of Metrics}
\label{app:metrics}
Evaluation for both EDH and TFD is done by SR (success rate), GC (goal condition success rate), and their path-length-weighted versions. Success Rate (SR) is a binary indicator of whether all subtasks were completed. The definition of ``subtasks'' is different for EDH and TfD; for the former, they are all tasks required to realize state changes done by the \textbf{human demonstration} that are relevant to the ultimate task (e.g. The demo state changes in each session of Fig.\ref{fig:demo} (b)). Thus, the state changes brought by the human is considered ground truth in EDH evaluation; this brings multiple challenges further discussed in \S\ref{sec:eval}. On the other hand, for TfD, the subtasks are independent of what was done in the demo; for example, as long as an agent ``slices the tomato'' correctly for the task of Fig.\ref{fig:demo} (b), its SR will be 1 for this task. \footnote{While the github repository \url{https://github.com/alexa/teach\#downloading-the-dataset} 
mentions that the EDH tasks were filtered so that ``the state changes checked for to evaluate success are only those that contribute towards task success in the main task of the gameplay session the EDH instance is created from'', we find that even after this filtering, there exist many EDH tasks with subotpimal demonstrations as in Fig.\ref{fig:demo}.}

\Tiffanyarxiv{The goal-condition success (GC) of a model is the ratio of goal-conditions completed at the end of an episode. Both SR and GC can be weighted by (path length of the expert trajectory)/ (path length taken by the agent); these are called path length weighted SR (PLWSR) and path length weighted GC (PLWGC). Higher is better for all metrics.}

\subsection{Challenges in Evaluation}
\label{sec:eval}

\paragraph{Irrelevant Actions} Humans often explore the environment, or simply play around in the middle of a task. This means they may flip a switch completely unrelated to the goal. Table \ref{tab:tab1} are representative state changes \Tiffanyarxiv{that do \textit{not} have direct correspondence with the dialogue, and the percentage of human demonstrations that contain these actions}.

\begin{table}
\begin{center}
    \begin{footnotesize}
    \noindent 
    \begin{tabular}{lcc}
          \toprule
          \multicolumn{1}{@{}l}{\textbf{Unnecessary State Changes}} & \textbf{Val Seen} & \textbf{Val Unseen} \\
         \toprule
          Coffee Machine on/off & 47.73  & 47.54 \\
          Picked up and not placed & 25.49  & 23.41 \\
          Faucet on/off & 12.68 & 10.59  \\
          Stove/ Microwave on/off & 35.61  & 28.31  \\
          
          $\Rightarrow$ Total & 38.98 & 35.60 \\
         \bottomrule
    \end{tabular}
    \end{footnotesize}
\end{center}
\caption{Representative state changes that do not have direct correspondence with the dialogue, and the percentage of human demonstrations that contain these actions. The action types listed here bring ``state changes'' that are counted during EDH evaluation. For example, an agent would ``fail'' an EDH task if the human annotator of the task left coffee machine off at the end, although the task (e.g. ``Make coffee'') or dialogue itself does not mention that it be left on.}
\label{tab:tab1}
\end{table}
  
It is not always clear if this behavior is because of misunderstandings, boredom\Tiffanyarxiv{, or curiosity}. For example, we can classify a large number of navigation and interaction "No Op"s, or action sequences that return to the original state (e.g. turning around in place). In principle, these might be information seeking, to build a better map of the environment, but in practice, \Tiffanyarxiv{ many of the demonstrations} do not seem to exhibit those properties, particularly in extreme cases like repeatedly picking up and putting down the same object. The percentages of prevalence of these unnecessary actions in both training and validation are shown in Table ~\ref{tab:tab2}.

\begin{table}[!h]
\begin{center}
\begin{footnotesize}
    \noindent \begin{tabular}{@{}@{\hspace{3pt}} p{12em} rrr@{}}
         \toprule
          \multirow{1}{*}{\textbf{Suboptimal Actions}} & \textbf{Train}  & \textbf{ V. Seen} &  \textbf{V. Un}    \\
         \toprule
         \multicolumn{4}{@{}l}{\textbf{Navigation No Op}} \\
          Turn Left/ Right x 4 & 3.07 & 2.30 & 2.33      \\
          Forward + Backward & 4.80 & 7.57 & 4.23  \\
          Pan Right + Pan Left & 5.83 & 4.77 & 8.10      \\
          Turn Right + Turn Left & 13.13 & 13.49 &  10.89   \\
          $\Rightarrow$ Total  &  22.41 & 23.03 & 21.36  \\
         \midrule
         \multicolumn{4}{@{}l}{\textbf{Interaction No Op}}\\
          Toggle off + on same obj & 1.39 & 1.81 & 1.58  \\
          Open + Close same obj & 1.46 & 1.80 & 1.30\\
          Place + Pick up same obj  & 25.06 & 27.80 & 28.34   \\
          $\Rightarrow$ Total  & 26.76 & 30.10 & 30.57   \\
          \midrule
         \multicolumn{1}{@{}l}{\textbf{Interaction w. unrelated obj.}} & 14.10  & 16.61  & 13.59 \\
          \midrule
         \multicolumn{1}{@{}l}{\textbf{Demo unaligned w. dialog}} & 25.25  & 22.49  & 23.40  \\ 
         \bottomrule
    \end{tabular}
\end{footnotesize}
\end{center}
\caption{Representative unnecessary action types that do not have associations with the high level goal or the dialogue, and the percentage of demonstrations that contain these action types in train/ valid seen/ valid unseen splits.}
\label{tab:tab2}
\end{table}

The prevalence of these actions can be viewed as a positive for realism and even helpful if teaching how to search, but pose a challenge for evaluation.  

\paragraph{Penalizing Agents for Accuracy}
Using a human's action trace as the ground truth, means agents are penalized for skipping erroneous actions.  This leads to a misleading mismatch in performance between EDH and TfD. Additionally, EDH inflates model performance as it includes subsequences which are nearly deterministic (e.g. all but the last ``placing" action). Table \ref{tab:tab3} contains EDH scores for our three comparison models and TfD for ET/FILM. As suggested by authors of related papers, we treat Unseen Success Rate as the most important metric (seen in \textcolor{blue}{\textbf{blue}}).

\begin{table}
\begin{center}
\begin{footnotesize}
\begin{tabular}{@{\hspace{3pt}}l@{\hspace{10pt}}r@{\hspace{5pt}}r@{\hspace{4pt}}rr@{\hspace{5pt}}r@{}}

\toprule
\hspace{0.5em} \textbf{}  & \multicolumn{2}{c}{\textbf{Valid Seen}} && \multicolumn{2}{c}{\textbf{Valid Unseen}} \rowsqueeze \\
 \cmidrule{2-3}\cmidrule{5-6}\rowsqueeze
  & \multicolumn{1}{c}{GC}  & \multicolumn{1}{c}{SR} && \multicolumn{1}{c}{GC}  &\multicolumn{1}{c}{\textbf{\textcolor{blue}{SR}}} \\
\midrule
\multicolumn{6}{@{}l}{\textbf{Entire Dialogue History }(\textbf{EDH})} \\
\textsc{E.T.} & 
               15.7 [4.1] &  10.2 [1.7] && 9.1[1.7] & 7.8[0.9]  \\

\textsc{Symbiote}  &
               25.9 [5.3]   &    \textbf{16.1 [2.6]} && 17.2 [2.9]  & 10.1 [1.2] \\  
\textsc{FILM}  &
               \textbf{26.4 [5.6]}   &    14.3 [2.1] && \textbf{18.3 [2.7]}   &\textcolor{blue}{\textbf{10.2 [1.0]}}\\   
\midrule
\multicolumn{6}{@{}l}{\textbf{Trajectory from Dialogue }(\textbf{TfD})} \\
\textsc{E.T.}      
             & 1.4 [4.8] &  1.0 [0.2]  && 0.4 [0.6] &  0.5 [0.1]  \\
\textsc{FILM} &  
            \textbf{5.8 [11.6]}  & \textbf{5.5 [2.6]}   && \textbf{6.1 [2.5]}  &  \textcolor{blue}{\textbf{2.9 [1.0]}}\\  
\bottomrule
\end{tabular}
\end{footnotesize}
\end{center}
\caption{EDH and TfD performances of E.T., Symbiote, and FILM. While the SR on TfD is very low for all models, E.T.'s performance on TfD drops significnatly due to replication of errors and lack of grounding of high-level semantics.}
\label{tab:tab3}
\end{table}

Note, that an ideal evaluation would capture both ``actions in context" and ``task success." In the following section breakdown the overall numbers presented here to understand if models more carefully.

\subsection{Challenges in Training}

\paragraph{Behavior Cloning with Suboptimal Demonstrations}

We find that ET trained with behavior cloning repeats the same mistakes in novel scenes that are frequent in demonstrations. We examine two kinds of mistakes in demonstrations - (1) No Op interactions, in which consecutive interactions produces futile state changes (e.g. Placing and immediately picking up the same object) and (2) Interactions with unrelated objects (e.g. picking up ``saltshaker'' while making coffee).  
In Table \ref{tab:tab4} we compare what percent of model predictions in seen and unseen scenes replicate the no-op behavior.

\begin{table}[!h]
\begin{center}
\centering
\footnotesize

\setlength\tabcolsep{2pt}
    \centering
    \begin{tabular}{@{} p{10em} rr r rr r rr@{}}
    \toprule
          \multirow{2}{*}{\textbf{Suboptimal Actions}} &
          \multicolumn{2}{c}{\textbf{ET}} & & 
          \multicolumn{2}{c}{\textbf{\katerina{}}}  & & 
          \multicolumn{2}{c}{\textbf{FILM}}   \\
          \cmidrule{2-3} \cmidrule{5-6}  \cmidrule{8-9}  
          & \multicolumn{1}{c}{S} & \multicolumn{1}{c}{U} & & \multicolumn{1}{c}{S} & \multicolumn{1}{c}{U} & &  \multicolumn{1}{c}{S} & \multicolumn{1}{c}{U}\\ 
         \toprule
         \multicolumn{9}{@{}l}{\textbf{No Op (same obj)}}\\
          Toggle off + on        &  0.0  & 0.1  && 0.2  & 0.1 && 0.0   & 0.0    \\
          Open + Close           &  2.5  &  1.5 && 0.0  & 0.2 && 0.0   & 0.0    \\
          Place + Pick up        & 45.1  & 47.1 && 4.9  & 2.5 && 0.0   & 0.0    \\
          $\Rightarrow$ Total    & 46.2  & 48.1 && 5.1  & 2.8 && 0.0   & 0.0    \\
          \midrule                                 
         \textbf{Unrelated obj.} & 24.0  & 20.3 && 27.5 & 30.6&& 15.9  &  12.10 \\
         \bottomrule
   
    \end{tabular}

\end{center}
\caption{Percentage of tasks in which a model exhibited replication of No Op actions.}
\label{tab:tab4}
\end{table}

\Tiffanyarxiv{While hard to quantify, we also note that the higher intention of seemingly unnecessary human demonstrations (e.g. to explore, to understand, etc) are not replicated by ET. This is backed by our observation that ET tends to be stuck in many (10 or more) repetitions of the same No Op/ unnecessary actions, until the end of the task or before resuming to perform other actions.}

Note that even \katerina{} is exhibiting some no-op behaviors, but as the model supervision/structure becomes more abstract (ET vs FILM) this disappears, leaving only object choice errors.

\paragraph{Grounding Queries}

Key to dialogue is language based information seeking.
A target object may be located in a closed receptacle (cabinet, etc); in this case, the agent has to query the commander for its location, as a human would. We examine whether models ground \textit{query utterances} into meaning and accurate actions, since this is one essential aspect of dialog grounding. While there are utterances with other essential intents, such as confirmation, we focus on query utterances since these are relatively easy to extract mechanically. 

\Tiffanyarxiv{In Table \ref{tab:tab5} we consider a subset of tasks that involve ``query utterances'' that can be detected automatically. Specifically, we present the performance of \Tiffany{models} in terms of success rate and goal condition success on tasks that require opening a receptacle based on the answer to a question -- and then measure if the models leverage the query. Not all query utterances will be of this type, but these tasks necessarily involve grounding query utterances for task success.}

\begin{table}[t]
\begin{center}
\footnotesize

\begin{tabular}{@{\hspace{5pt}}l@{\hspace{15pt}}cccc@{}}
            
\toprule   
Method     & SR & GC & SR w. Query & GC w. Query \\
\toprule   
\multicolumn{3}{@{}l}{\textbf{Validation Seen}}\\

ET          & 8.97 & 14.13& 0.00   & 0.00 \\%
\katerina{} & 0.00 & 11.76& 0.00   & 0.00 \\%
FILM        & 9.59 & 20.26& 0.00   & 0.00 \\%

\midrule   
\multicolumn{3}{@{}l}{\textbf{Validation Unseen}}\\
ET              & 2.39 & 8.69 & 0.00  & \textbf{0.49}\\%
\katerina{}     & 0.00 & 0.00 & 0.00    & 0.00 \\%
FILM            & 1.29 & 9.79 & 0.00  & 0.00 \\%

\bottomrule
\end{tabular}

\end{center}
\caption{We consider a task as involving ``query utterances'', if in its demonstration, a relevant object inside an originally closed receptacle was picked up. SR/GC measure the vanilla task success on tasks with ``query utterances''; SR/GC w. Query measure if the success was achieved using information in the ``query utterances.''}
\label{tab:tab5}
\end{table}

Queries are present in 23.05\% and 25.31\% of valid seen and unseen splits, respectively.  This is a key challenge as it demonstrates a clear use case for dialogue and limitation of current models.

Given a statement like ``the fork is \textit{in the cabinet} left to the refrigerator'', the evaluation mismatch occurs if an agent grabs a different fork on a table. This allows them to succeed, as measured by  SR/GC, but not in SR/GC with Query. Notably, all models fail at query grounding, indicating they are simply ignoring the language instructions. This shows that enabling complex dialogue grounding is an important open problem for DIF. Especially, for the ultimate goal of two-agent task completion (TaTC), it is necessary that models can ground query and other essential utterances in a dialogue.

\section{Conclusion and Next Steps}
This paper is not an indictment of TEACh, nor an endorsement of a particular model, rather it seeks to lay out important questions and challenges that NLP will need to tackle as it moves into embodied dialogue.  Unlike existing work in dialogue that looks to model human satisfaction \cite{NEURIPS2019_fc981212} or state-tracking, DIF has the advantage of explicit and verifiable semantic goals.  We pose a challenge to the community: How can we build agents where success is not tied to specific actions yet language understanding and production are accurate and fluent? As a first step, we posit that imitation learning should be avoided.

\section{Limitations}
We focus on a new embodied benchmark -- there is substantial work in dialogue (including goal directed) for non-embodied environments which we do not consider as aligned with the goals of embodied DIF, but may have important insights. Additionally, future work may overcome the issues raised and it is unclear how to transfer our findings back to the goal directed dialogue in the non-embodied setting. Additional insights may derive from research in social intelligence. 

\bibliography{anthology,custom}
\bibliographystyle{acl_natbib}

\clearpage
\appendix

\section{More Discussion of Symbiote}
\label{app:symbiote}
Symbiote has a modular structure, which consists of language understanding, mapping, and low-level planning components. It is not trained with imitation learning of low-level demonstrations (e.g. move right, move left, etc.). Demonstrations are used only in the sense that they provide subgoals that suervise the training of the language understanding component. 

More specifically, a pretrained T5 model \cite{T5} fine-tuned with the ground truth subgoals (\texttt{edh\_instance[`future\_subgoals']}), serves as the language understanding component. The model takes the driver and commander's dialogue and previous actions as input; it is trained to output a sequence of subgoals of the form ``\{action\} \{obj\}'', where \{action\} is either ``navigate'' or any of the primitive interactions commands "pickup", "cut", "toggle", etc, and \{obj\} is any of the object classes in ai2thor. 

For the mapping component, a DETR detector \cite{DETR} was finetuned on the train set scenes of TEACh and the depth prediction model from FILM was used off-the-shelf. Frontier based exploration is used for environment exploration. Similarly as in FILM, the agent navigates to object goals in the map using the fast marching method.

\section{How the Statistics of Section 5 were Obtained}
\label{app:stats}

We explain how the statistics that appear in each table of Section 4 were obtained. All analyses, except for TfD results in \textbf{Penalizing Agents for Accuracy}, were done on EDH tasks. 

\paragraph{Irrelevant Actions}

The first table shows some representative unnecessary state changes that EDH tasks require for ``task success' in evaluation. For example, in our common sense, it is not necessary that we leave the coffee machine on to successfully make coffee (indeed, it is better to turn it off after use). However, since EDH evaluation requires that the agent exactly follows state changes done in the demonstration, the agent will have to leave coffee machine turned on for a particular validation task, if this was done in its corresponding demonstration. 

Each row shows unnecessary state changes that are exemplary and the average frequency of these noises across relevant tasks. More specifically, 

\begin{itemize}
    \item \textbf{Coffee Machine on/ off}: \textit{`Coffee'} tasks
    \item \textbf{Picked up and not placed}: all tasks
    \item \textbf{Faucet on/ off}: all tasks that may involve using the faucet (\textit{`Coffee', `Clean All X', `Boil X',`Water Plant',`Sandwich', `Breakfast', `Plate Of Toast', `Salad'})
    \item \textbf{Stove/ Microwave on/off}: all tasks that may involve using a heating appliance (\textit{`Boil X',`N Cooked Slices Of X In Y'})
\end{itemize}

``Total'' accounts for the percentage of EDH tasks that fall into any of the above criteria. Please refer to \cite{TEACH} for the possible types (e.g. \textit{`Coffee'}) of tasks.

While the first table shows statistics of irrelevant state changes of ``relevant objects'', the second table shows those of more random actions, at a lower level. Navigation No Op, the first kind, was simply obtained by detecting the existence of consecutive Turn Lef/Right x 4, Forward + Backward, Pan Right + Pan Left, Turn Right + Turn Left. The second kind, interaction No Op, was similarly detected. Whether an consecutive and opposite interactions were done on the same ``object'' was detected by replaying the \texttt{pred\_actions} in the model outputs. Interaction w. unrelated objects denotes whether the demonstration an object that is completely unrelated from task type (e.g. picking up \textit{saltshaker} for a task whose type is \textit{`Coffee'}). Demonstrations unaligned with dialogue were counted manually since there is no automatic way to filter these.

\paragraph{Penalizing Agents for Accuracy}

The statistics in this subsection were straightforwardly obtained by averaging over the evaluation outputs (whose formats follow that of the original ET code from TEACh) of each task. 

\paragraph{Behavior Cloning with Suboptimal Demonstrations}

The same procedures for the second table in \textbf{Irrelevant Actions} were used.

\section{TEACh Prefiltering}
\label{app:filter}
Only necessary state changes are checked in EDH evaluation, but all are present in training.   \url{https://github.com/alexa/teach#downloading-the-dataset} mentions that the authors filtered the EDH tasks so that ``the state changes checked for to evaluate success are only those that contribute towards task success in the main task of the gameplay session the EDH instance is created from.'' Our analysis is on data that has already been filtered and cleaned and yet still exhibits these problems. 

\label{sec:appendix}

\end{document}